\documentclass[runningheads]{llncs}
\usepackage[T1]{fontenc}
\usepackage{graphicx}
\usepackage{amsmath}
\usepackage{amssymb}
\usepackage{booktabs}
\usepackage{hyperref}
\usepackage{url}
\usepackage{subcaption}

\begin{document}

\title{SmolVLA-MoE: Sparse Mixture-of-Experts for Efficient Multi-Task Robotic Manipulation}

\titlerunning{SmolVLA-MoE}

\author{Shiyang Zhang\inst{1} \and Miao Yang\inst{1} \and Zhengda Li\inst{2} \and Yang Shao\inst{2} \and Ruifang Zhai\inst{1,3} \and Hui Peng\inst{1}\thanks{Corresponding author} \and Peng Song\inst{2,3}\thanks{Corresponding author}}

\authorrunning{S. Zhang et al.}

\institute{College of Informatics, Huazhong Agricultural University, No. 1 Shizi Mountain Street, Hongshan District, Wuhan 430070, China
\and
College of Plant Science and Technology, Huazhong Agricultural University, No. 1 Shizi Mountain Street, Hongshan District, Wuhan 430070, China
\and
Hubei Hongshan Laboratory, Wuhan 430070, China\\
\email{2305304089@qq.com, moonbird@mail.hzau.edu.cn, songp@mail.hzau.edu.cn}}

\maketitle

\begin{abstract}
Vision-Language-Action (VLA) models have shown promising results in multi-task robotic manipulation by leveraging visual and language understanding for action generation. However, existing lightweight VLA models employ a single dense action expert to handle all tasks, which limits their multi-task capability under constrained computational budgets. To address this issue, we propose SmolVLA-MoE, a novel framework that introduces sparse Mixture-of-Experts (MoE) into the FFN layers of the action expert in VLA models. SmolVLA-MoE replaces dense FFN layers with multiple parallel expert FFNs and a token-level router, enabling each token to adaptively select the most suitable expert for action generation. We further introduce a shared expert with a tunable weight parameter to balance the inheritance of pre-trained capabilities and task-specific expert specialization. Through sparse activation, SmolVLA-MoE activates only a single expert FFN ($\sim$100M activated parameters in the action expert module) during inference, with a total activated parameter count of $\sim$500M including the frozen VLM backbone---the same as the dense baseline SmolVLA---while benefiting from the multi-expert representational capacity provided by $\sim$800M total stored parameters. We evaluate our method on four task suites of the LIBERO benchmark and compare it against SmolVLA and other representative baselines. Experimental results demonstrate that SmolVLA-MoE achieves an average success rate of 84.25\% across all four suites, outperforming the dense baseline SmolVLA (82.75\%) and the much larger OpenVLA-7B (76.5\%), while maintaining comparable inference efficiency, providing an effective and efficient solution for multi-task robotic manipulation. Our code is publicly available at \url{https://github.com/dudu435/smolvla-moe}.

\end{abstract}

\section{Introduction}
\label{sec:introduction}

Robotic manipulation in unstructured environments requires models that can understand diverse visual scenes, comprehend natural language instructions, and generate corresponding actions. Vision-Language-Action (VLA) models~\cite{brohan2023rt2,brohan2023rt1,octo2024,kim2024openvla} have emerged as a promising paradigm to address this challenge by unifying vision, language, and action within a single framework. Recent works such as RT-1~\cite{brohan2023rt1}, RT-2~\cite{brohan2023rt2}, Octo~\cite{octo2024}, and OpenVLA~\cite{kim2024openvla} have demonstrated that VLA models can leverage large-scale pre-trained Vision-Language Models (VLMs)~\cite{liu2023llava,beyer2024paligemma} as backbones to effectively generalize across various manipulation tasks.

Despite their strong performance, most VLA models rely on large-scale architectures with billions of parameters, posing significant challenges for real-time deployment on resource-constrained robotic platforms. To address this limitation, lightweight VLA models such as SmolVLA~\cite{smolvla2025} have been proposed, which employ a compact VLM backbone (e.g., SmolVLM2-500M) combined with a small action expert network for action generation via Flow Matching. This design significantly reduces computational cost while maintaining competitive performance.

However, these lightweight VLA models suffer from a fundamental limitation: the action expert uses a single dense feed-forward network (FFN) to uniformly process all tasks. In multi-task robotic manipulation scenarios, different tasks (e.g., spatial reasoning, object manipulation, goal reaching) exhibit distinct action patterns that require different computational strategies. A single dense FFN is forced to encode all these diverse action generation patterns into a shared set of parameters, creating an inherent tension between model capacity and computational efficiency. Simply scaling up the action expert proportionally increases inference cost, negating the advantages of the lightweight design.

Mixture-of-Experts (MoE)~\cite{jacobs1991adaptive,shazeer2017moe,jiang2024mixtral,deepseek2024moe} offers an elegant solution to this dilemma. Unlike model compression techniques that trade representational capacity for efficiency, sparse MoE scales model capacity by maintaining multiple parallel expert networks while routing each input to only a sparse subset, keeping the active computation constant. This paradigm has proven effective in large language models such as Mixtral~\cite{jiang2024mixtral} and DeepSeek-MoE~\cite{deepseek2024moe}, yet its application in VLA models for action generation remains largely unexplored.

In this paper, we propose SmolVLA-MoE, a framework that introduces sparse Mixture-of-Experts into the FFN layers of the action expert in VLA models. Unlike approaches that replicate the entire action expert into multiple copies, our method operates at a finer granularity, replacing each dense FFN layer with multiple parallel expert FFNs and a learnable token-level router. This design enables each token to adaptively select the most suitable expert for action generation based on the input context, while keeping the attention layers shared across all tasks.

We evaluate SmolVLA-MoE on the LIBERO benchmark~\cite{liu2023libero}, which provides four task suites (Spatial, Object, Goal, and LIBERO-10) covering diverse manipulation challenges. Experimental results show that SmolVLA-MoE achieves superior performance over the SmolVLA baseline on all four suites, while activating only a single expert FFN ($\sim$100M activated parameters in the action expert) during inference, keeping the total activated parameter count at $\sim$500M---the same as the dense baseline. We further conduct systematic ablation studies on key design parameters, including the number of activated experts (top-$k$) and the shared expert weight, providing in-depth analysis of MoE behavior in robotic action generation.

Our main contributions are summarized as follows:

\begin{itemize}
  \item We propose SmolVLA-MoE, the first work to build a Mixture-of-Experts model on top of the lightweight SmolVLA architecture, maintaining the lightweight and efficient characteristics of SmolVLA under the same activated parameter budget.
  \item Building upon the AdaMoE~\cite{zeng2024adamoe} design, we introduce more controllable weight gating between shared and routed experts to balance pre-trained capability inheritance and expert specialization, and conduct systematic analysis of its impact across different task types.
  \item We demonstrate that SmolVLA-MoE achieves state-of-the-art performance across all four task suites of the LIBERO benchmark while maintaining inference efficiency comparable to the dense baseline.
\end{itemize}

\section{Related Work}
\label{sec:related_work}

\subsection{Vision-Language-Action Models}

Vision-Language-Action (VLA) models aim to leverage pre-trained vision-language models for action generation, building general-purpose robotic policies. RT-1~\cite{brohan2023rt1} first demonstrated that a Transformer-based model trained on large-scale real-world robotic data can generalize across hundreds of tasks. RT-2~\cite{brohan2023rt2} pioneered the VLA direction by fine-tuning a large vision-language model (55B-parameter PaLI-X) to directly output robotic actions as text tokens, demonstrating that semantic knowledge embedded in VLMs can transfer to robotic control. Octo~\cite{octo2024} proposed a Transformer-based policy architecture supporting flexible task and observation specifications, enabling multi-task learning across multiple robot embodiments. OpenVLA~\cite{kim2024openvla} built a 7B-parameter open-source VLA model based on the Llama 2 architecture, achieving strong generalization through large-scale pre-training on the Open X-Embodiment dataset~\cite{openx2024}.

However, these models typically require billions of parameters, making real-time deployment on robotic platforms challenging. Meanwhile, alternative policy learning approaches such as ACT~\cite{zhao2023act} and Diffusion Policy~\cite{chi2023diffusion_policy} have shown strong performance in imitation learning with more compact architectures, and vision-language foundation models have been adapted for robotic control~\cite{li2023roboflamingo}. To this end, recent works have explored lightweight VLA architectures. $\pi_0$~\cite{black2024pi0} proposed a flow matching-based VLA that uses a pre-trained vision-language model (PaliGemma~\cite{beyer2024paligemma}) as the backbone and trains a dedicated action expert for action chunk generation. SmolVLA~\cite{smolvla2025} adopts a compact VLM backbone (SmolVLM2-500M) and introduces an independent action expert network that generates actions via Flow Matching~\cite{lipman2023flow}, significantly reducing model size while maintaining competitive performance. This lightweight approach demonstrates that effective robotic manipulation does not necessarily require billion-scale models, but its single dense action expert design limits multi-task capability.

Our work builds upon SmolVLA by introducing sparse MoE into the action expert to address this limitation, enabling efficient multi-task scaling without increasing inference cost.

\subsection{Mixture-of-Experts Models}

Mixture-of-Experts (MoE)~\cite{jacobs1991adaptive} is a classical approach for scaling model capacity through conditional computation, maintaining multiple expert networks and using a gating mechanism to route each input to a sparse subset of experts. The sparsely-gated MoE layer~\cite{shazeer2017moe} first demonstrated the scalability of this approach in deep learning. Subsequent works have advanced MoE along several directions, including simplified top-1 routing with load balancing~\cite{fedus2022switch} and large-scale top-2 routing~\cite{lepikhin2021gshard}. More recently, Mixtral~\cite{jiang2024mixtral} demonstrated that sparse MoE can match much larger dense models at a fraction of the inference cost. Two designs are particularly relevant to our work: DeepSeek-MoE~\cite{deepseek2024moe} introduced shared experts that are always activated alongside routed experts, providing a stable baseline; AdaMoE~\cite{zeng2024adamoe} proposed adaptive scaling adapters that decouple expert selection from expert weighting.

In the robotics domain, the application of MoE remains limited. Existing approaches to efficient robot learning primarily focus on model compression: knowledge distillation~\cite{hinton2015distilling} transfers knowledge from large teacher models to smaller students, parameter-efficient fine-tuning methods like LoRA~\cite{hu2022lora} reduce trainable parameters, and data augmentation techniques~\cite{yu2023scaling,mandlekar2023mimicgen} improve sample efficiency. However, these methods fundamentally trade off representational capacity for computational savings. Sparse MoE offers a distinct advantage: it scales model capacity through multiple specialized experts while keeping the per-inference computation constant, making it ideally suited for multi-task robotic manipulation where different tasks demand different computational strategies but the deployment budget is fixed. Some works have explored MoE in multi-task reinforcement learning~\cite{willi2024moe_rl}, but the integration of MoE into lightweight VLA models for action generation has not been systematically studied. To this end, our work introduces FFN-level MoE into the action expert of VLA models, drawing on the shared expert design of DeepSeek-MoE~\cite{deepseek2024moe} and the adaptive routing of AdaMoE~\cite{zeng2024adamoe}, combined with Sparse Upcycling~\cite{komatsuzaki2023sparse} initialization that converts pre-trained dense checkpoints into sparse MoE models while preserving pre-trained knowledge.

\section{Method}
\label{sec:method}

\begin{figure}[t]
  \centering
  \includegraphics[width=0.95\textwidth]{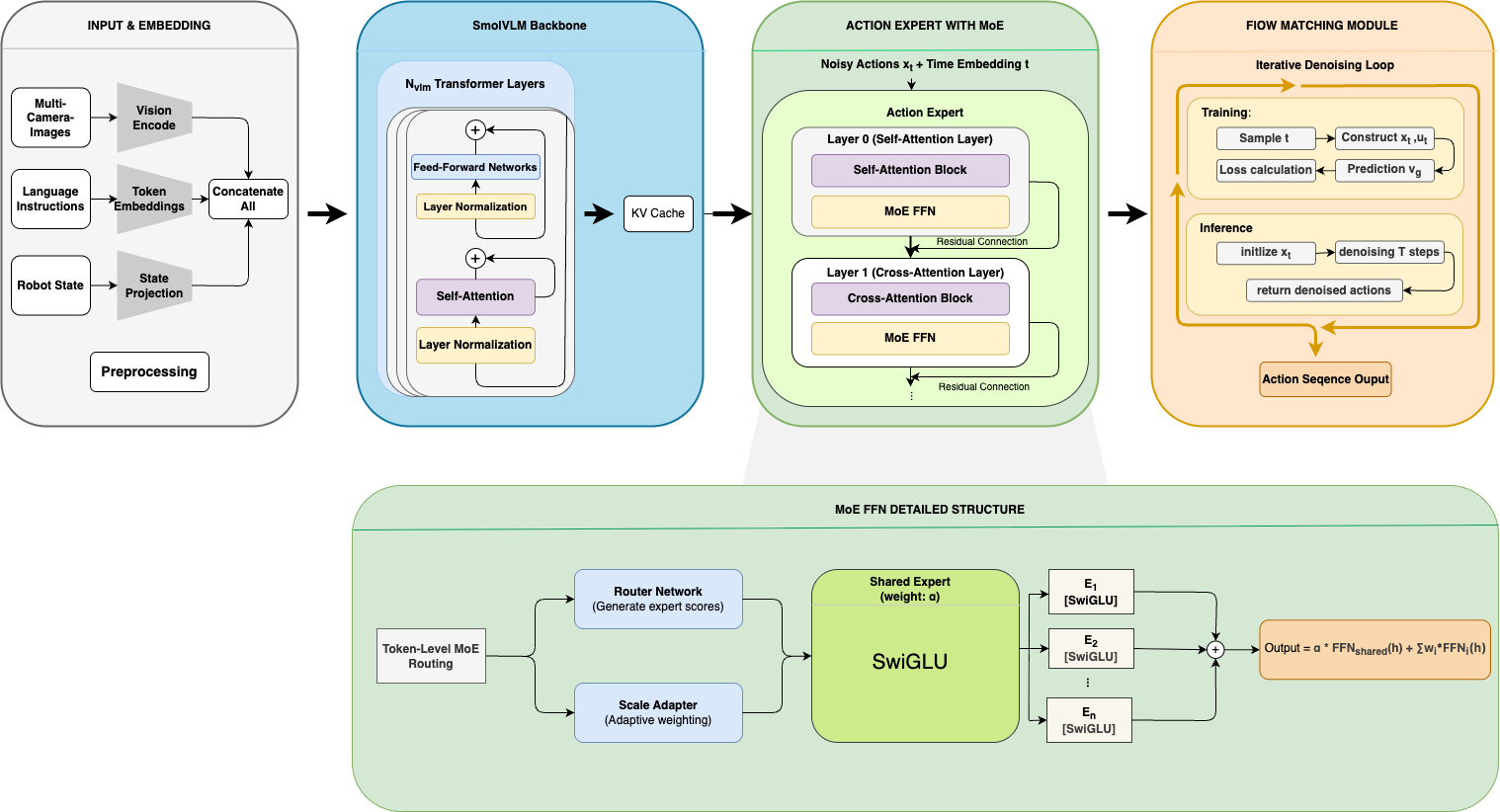}
  \caption{Overview of the SmolVLA-MoE architecture. The frozen VLM backbone processes visual observations and language instructions to produce contextual representations. The action expert replaces dense FFN layers with MoE-FFN layers, each containing a shared expert $E_s$ (always activated, weighted by $\alpha$), $N$ routed expert FFNs, a token-level router with top-$K$ selection, and a scaling adapter. Through sparse activation, only $K$ experts are selected per token during both training and inference.}
  \label{fig:architecture}
\end{figure}

This section presents SmolVLA-MoE in detail. We first review the SmolVLA architecture that serves as our baseline (Sect.~\ref{sec:preliminaries}), then describe the FFN-level Mixture-of-Experts design (Sect.~\ref{sec:moe_ffn}), the routing mechanism (Sect.~\ref{sec:routing}), the shared expert with tunable weight (Sect.~\ref{sec:shared_expert}), and the training objective (Sect.~\ref{sec:training}). The overall architecture is illustrated in Fig.~\ref{fig:architecture}.

\subsection{Preliminaries: SmolVLA}
\label{sec:preliminaries}

SmolVLA is a lightweight Vision-Language-Action model consisting of two main components: a frozen VLM backbone and a trainable action expert.

\textbf{VLM Backbone.} SmolVLA adopts SmolVLM2-500M as its vision-language backbone, comprising a SigLIP vision encoder and a language model. The VLM processes visual observations and language instructions to produce contextual representations. During training, the VLM backbone is frozen to preserve its pre-trained knowledge.

\textbf{Action Expert.} The action expert is a small Transformer network that generates robotic actions based on the VLM's representations. It consists of multiple Transformer layers, each containing a cross-attention module (attending to the VLM's key-value cache) and a feed-forward network (FFN). The FFN in each layer adopts the SwiGLU architecture:
\begin{equation}
  \text{FFN}(x) = W_{down} \cdot (\text{SiLU}(W_{gate} \cdot x) \odot W_{up} \cdot x)
\end{equation}
where $W_{gate} \in \mathbb{R}^{d_{ff} \times d}$, $W_{up} \in \mathbb{R}^{d_{ff} \times d}$, $W_{down} \in \mathbb{R}^{d \times d_{ff}}$ are learnable weight matrices, $d$ is the hidden dimension, and $d_{ff}$ is the intermediate dimension.

\textbf{Flow Matching.} SmolVLA uses Flow Matching~\cite{lipman2023flow,tong2023cfm} for action generation. Unlike diffusion-based approaches~\cite{ho2020ddpm,chi2023diffusion_policy} that require many denoising steps, Flow Matching learns a continuous velocity field along a straight-line transport path from noise to the target action distribution. During training, given a ground-truth action sequence $a \in \mathbb{R}^{H \times d_a}$ (where $H$ is the action chunk size and $d_a$ is the action dimension) and sampled noise $\epsilon \sim \mathcal{N}(0, I)$, a noisy action is constructed via linear interpolation at timestep $t \sim \text{Beta}(1.5, 1.0)$:
\begin{equation}
  x_t = t \cdot \epsilon + (1 - t) \cdot a
\end{equation}
where $t \in [0, 1]$ controls the interpolation between the clean action $a$ (at $t=0$) and pure noise $\epsilon$ (at $t=1$). The model learns to predict the velocity field $v_\theta(x_t, t)$ that transports $x_t$ toward $a$, with the target velocity $u_t = \epsilon - a$ representing the constant velocity along the straight-line path. The training loss minimizes the mean squared error between predicted and target velocities:
\begin{equation}
  \mathcal{L}_{FM} = \mathbb{E}_{t, \epsilon} \left[ \| v_\theta(x_t, t) - u_t \|^2 \right]
\end{equation}
At inference, actions are generated by integrating the learned velocity field backward from $t=1$ (pure noise) to $t=0$ (clean action) through $N$ steps of Euler integration with step size $\Delta t = -1/N$.

\subsection{FFN-Level Mixture-of-Experts}
\label{sec:moe_ffn}

The core idea of SmolVLA-MoE is to replace the dense FFN layers in the action expert with Mixture-of-Experts FFN (MoE-FFN) layers. Unlike full expert-level MoE that replicates the entire action expert into multiple copies, our method operates at the FFN granularity, keeping the attention layers shared while allowing each token to select specialized FFN experts.

\textbf{MoE-FFN Layer.} Each MoE-FFN layer contains $N$ parallel expert FFNs $\{E_1, E_2, \ldots, E_N\}$, a shared expert $E_s$, a router $R$, and a scaling adapter $S$. Each expert FFN adopts the same SwiGLU architecture as the original dense FFN but with independent parameters:
\begin{equation}
  E_i(x) = W_{down}^{(i)} \cdot (\text{SiLU}(W_{gate}^{(i)} \cdot x) \odot W_{up}^{(i)} \cdot x), \quad i \in \{1, \ldots, N\}
\end{equation}
where $W_{gate}^{(i)}$, $W_{up}^{(i)}$, and $W_{down}^{(i)}$ are the learnable weight matrices of the $i$-th expert, $x \in \mathbb{R}^d$ is the input token representation, $N$ is the total number of routed experts, and $\odot$ denotes element-wise multiplication. Each expert maintains its own independent set of parameters, allowing different experts to learn distinct transformation patterns.

\textbf{Layer Replacement.} The replacement frequency is controlled by a parameter $f$. When $f = 1$, all FFN layers are replaced with MoE-FFN; when $f = 2$, every other layer is replaced. For a layer with index $l$, MoE-FFN is applied when $l \bmod f = 0$.

\begin{figure}[t]
  \centering
  \includegraphics[width=0.7\textwidth]{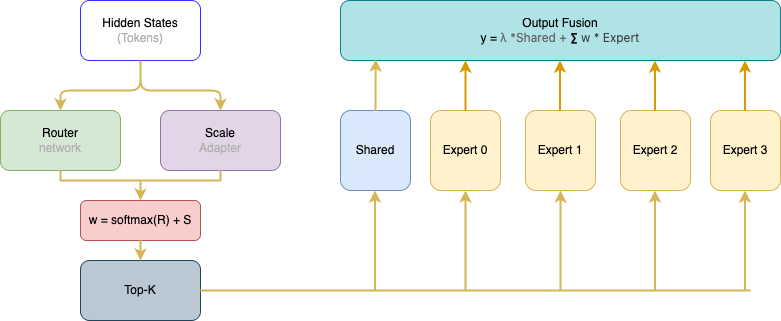}
  \caption{Detailed structure of the MoE-FFN layer. The input token is processed by both the shared expert and the top-$K$ routed experts selected by the router. The scaling adapter provides additional per-token adaptive scaling. The final output is a weighted combination of the shared expert output (scaled by $\alpha$) and the routed expert output (scaled by $1-\alpha$).}
  \label{fig:moe_ffn}
\end{figure}

\subsection{Routing Mechanism}
\label{sec:routing}

The routing mechanism determines which experts process each token. It consists of two components: the router and the scaling adapter.

\textbf{Router.} The router is a bias-free linear projection that maps each token's hidden state to expert scores:
\begin{equation}
  g(x) = W_r \cdot x, \quad W_r \in \mathbb{R}^{N \times d}
\end{equation}
where $g(x) \in \mathbb{R}^N$ contains the routing logits for $N$ experts.

\textbf{Top-K Selection.} For each token, the top $K$ experts with the highest routing logits are selected:
\begin{equation}
  \mathcal{T}_K = \text{TopK}(g(x), K)
\end{equation}
where $\mathcal{T}_K$ denotes the index set of the $K$ selected experts and $K$ is the number of experts activated per token. The routing weights are computed by applying softmax normalization to the logits of the selected experts, ensuring they sum to one:
\begin{equation}
  w_k = \text{Softmax}(g(x)_{\mathcal{T}_K})_k, \quad k \in \{1, \ldots, K\}
\end{equation}
where $g(x)_{\mathcal{T}_K}$ extracts the routing logits corresponding to the selected expert indices, and $w_k$ is the normalized routing weight for the $k$-th selected expert.

\textbf{Scaling Adapter.} Following AdaMoE~\cite{zeng2024adamoe}, we introduce a scaling adapter that provides additional per-token adaptive scaling, decoupling expert selection from expert weighting. The scaling adapter is also a bias-free linear projection:
\begin{equation}
  s(x) = W_s \cdot x, \quad W_s \in \mathbb{R}^{N \times d}
\end{equation}
where $W_s$ is the scaling adapter weight matrix, producing an $N$-dimensional scaling vector for each token. The final routing weights combine the router softmax weights with the scaling adapter outputs for the selected experts:
\begin{equation}
  \hat{w}_k = w_k + s(x)_{\mathcal{T}_K(k)}, \quad k \in \{1, \ldots, K\}
\end{equation}
where $\hat{w}_k$ is the adjusted routing weight for the $k$-th selected expert, and $s(x)_{\mathcal{T}_K(k)}$ is the scaling adapter output corresponding to that expert. This additive combination allows the scaling adapter to modulate the magnitude of expert contributions independently of the router's selection decision.

\textbf{Routed Output.} The routed expert output for each token is the weighted sum of the selected expert outputs:
\begin{equation}
  y_{routed}(x) = \sum_{k=1}^{K} \hat{w}_k \cdot E_{\mathcal{T}_K(k)}(x)
\end{equation}
where $E_{\mathcal{T}_K(k)}(x)$ is the output of the $k$-th selected expert applied to input $x$, and $\hat{w}_k$ is its corresponding adjusted routing weight. Only $K$ out of $N$ total experts are computed per token, achieving sparse activation.

\subsection{Shared Expert with Tunable Weight}
\label{sec:shared_expert}

Inspired by DeepSeek-MoE~\cite{deepseek2024moe}, we introduce a shared expert $E_s$ that is always activated for all tokens, providing a stable baseline output. The shared expert has the same SwiGLU architecture as the routed experts but with independent parameters.

The final output of the MoE-FFN layer combines the shared expert output and the routed expert output through a tunable weight parameter $\alpha$:
\begin{equation}
  y(x) = \alpha \cdot E_s(x) + y_{routed}(x)
\end{equation}
where $y(x)$ is the final MoE-FFN layer output, $E_s(x)$ is the shared expert output, $y_{routed}(x)$ is the aggregated routed expert output defined in Eq.~(14), and $\alpha \in [0, 1]$ is the shared expert weight that controls the relative contribution of the shared expert. This parameter provides continuous control between two complementary roles:

\begin{itemize}
  \item \textbf{Pre-trained capability inheritance} ($\alpha \to 1$): The shared expert preserves the pre-trained knowledge of the original dense FFN, ensuring training stability.
  \item \textbf{Expert specialization} ($\alpha \to 0$): The routed experts gain greater influence, enabling task-specific specialization.
\end{itemize}

Note that the shared expert is initialized with the exact pre-trained dense FFN weights without added noise, while the routed experts receive noise perturbation (Sect.~\ref{sec:training}). This asymmetric initialization ensures that the shared expert serves as a reliable baseline from the start of training.

\subsection{Training Objective}
\label{sec:training}

\textbf{Sparse Upcycling Initialization.} Starting from the pre-trained baseline, all expert FFNs are initialized from the pre-trained dense FFN weights of the original SmolVLA action expert. The shared expert receives an exact copy, while each routed expert receives the same weights with added Gaussian noise:
\begin{equation}
  W^{(i)} = W_{dense} + \sigma \cdot \text{std}(W_{dense}) \cdot \mathcal{N}(0, I), \quad i \in \{1, \ldots, N\}
\end{equation}
where $W^{(i)}$ is the initialized weight matrix of the $i$-th routed expert, $W_{dense}$ is the pre-trained dense FFN weight matrix, $\text{std}(W_{dense})$ is the standard deviation of $W_{dense}$ used to scale the noise relative to the weight magnitude, $\mathcal{N}(0, I)$ is standard Gaussian noise, and $\sigma$ is the noise scale (default 0.02). This initialization ensures: (1) the model starts from a strong pre-trained checkpoint, and (2) the noise promotes expert diversity and specialization during training.

\textbf{Load Balancing Loss.} Following Switch Transformer~\cite{fedus2022switch}, we adopt a load balancing loss to encourage uniform expert utilization and prevent expert collapse:
\begin{equation}
  \mathcal{L}_{LB} = N \cdot \sum_{i=1}^{N} f_i \cdot P_i
\end{equation}
where $N$ is the total number of routed experts, $f_i = \frac{1}{T}\sum_{j=1}^{T} \mathbf{1}[\text{expert } i \text{ is selected for token } j]$ is the fraction of tokens routed to expert $i$ across a batch of $T$ tokens (non-differentiable), and $P_i = \frac{1}{T}\sum_{j=1}^{T} p_i(x_j)$ is the average routing probability for expert $i$ computed from the softmax output (differentiable). The product $f_i \cdot P_i$ is minimized when both the actual token allocation and routing probabilities are uniformly distributed, thus penalizing imbalanced expert utilization.

\textbf{Total Loss.} The total training loss combines the Flow Matching loss and the load balancing loss:
\begin{equation}
  \mathcal{L} = \mathcal{L}_{FM} + \lambda \cdot \mathcal{L}_{LB}
\end{equation}
where $\mathcal{L}_{FM}$ is the Flow Matching loss defined in Eq.~(3), $\mathcal{L}_{LB}$ is the load balancing loss defined in Eq.~(17), and $\lambda$ is the load balancing weight (default 0.01) that controls the trade-off between action prediction accuracy and expert utilization balance. The load balancing loss is averaged over all MoE-FFN layers in the action expert.

\section{Experiments}
\label{sec:experiments}

This section evaluates SmolVLA-MoE on the LIBERO benchmark. We first describe the experimental setup (Sect.~\ref{sec:setup}), then present the main results (Sect.~\ref{sec:main_results}), followed by ablation studies (Sect.~\ref{sec:ablation}).

\subsection{Experimental Setup}
\label{sec:setup}

\textbf{Benchmark.} We conduct evaluations on the LIBERO benchmark~\cite{liu2023libero}, which provides standardized evaluation suites for multi-task robotic manipulation in simulated tabletop environments. LIBERO contains four task suites of increasing difficulty:

\begin{itemize}
  \item \textbf{LIBERO-Spatial}: 10 tasks focusing on spatial relationship understanding (e.g., ``pick up the object to the left of the bowl''). This suite tests the model's ability to comprehend spatial language instructions.
  \item \textbf{LIBERO-Object}: 10 tasks focusing on object recognition and manipulation (e.g., ``pick up the red cup''). This suite evaluates the model's ability to distinguish between different objects.
  \item \textbf{LIBERO-Goal}: 10 tasks focusing on goal-conditioned manipulation with diverse goals. This suite tests the model's ability to reach specified goal states.
  \item \textbf{LIBERO-10}: 10 tasks combining elements from all categories with multi-step instructions. This is the most challenging suite, requiring compositional understanding of spatial, object, and goal concepts.
\end{itemize}

Each task is evaluated over 10 rollouts, using success rate as the evaluation metric.

\begin{figure}[t]
  \centering
  \begin{subfigure}[b]{0.85\textwidth}
    \centering
    \includegraphics[width=\textwidth]{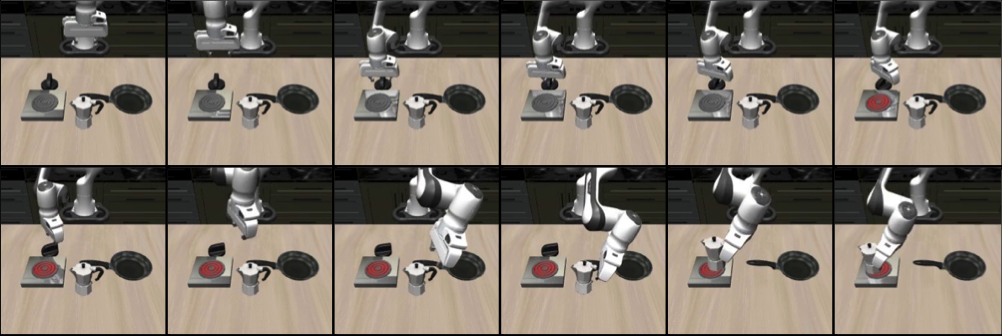}
    \caption{``Turn on the stove and put the moka pot on it.''}
    \label{fig:task_9}
  \end{subfigure}
  \vspace{4pt}
  \begin{subfigure}[b]{0.85\textwidth}
    \centering
    \includegraphics[width=\textwidth]{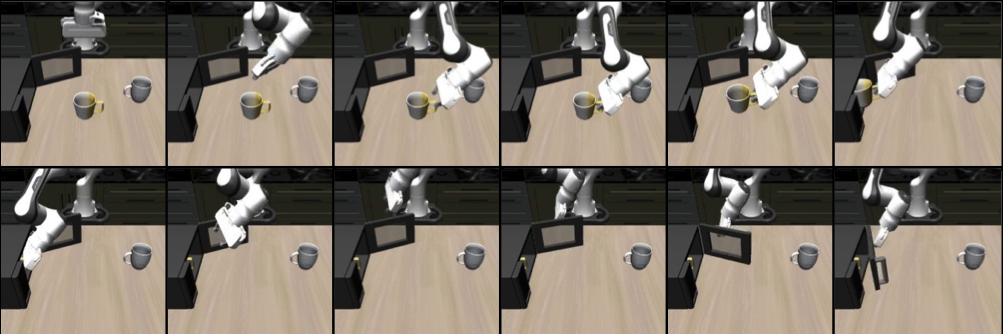}
    \caption{``Put the yellow and white mug in the microwave and close it.''}
    \label{fig:task_7}
  \end{subfigure}
  \caption{Visualization of complete rollouts in the LIBERO simulation environment. Each row shows key frames of the execution trajectory for a multi-step manipulation task from start to completion.}
  \label{fig:task_demo}
\end{figure}

\textbf{Baselines.} We compare SmolVLA-MoE against six representative baselines spanning different scales and architectures: Diffusion Policy~\cite{chi2023diffusion_policy} (321M), Octo~\cite{octo2024} (93M), OpenVLA~\cite{kim2024openvla} (7B, pre-trained on Open X-Embodiment~\cite{openx2024}), $\pi_0$ in two configurations~\cite{black2024pi0} (Paligemma-3B and 3.3B), and SmolVLA~\cite{smolvla2025} (500M) as the direct dense baseline fine-tuned on each LIBERO suite following the same protocol.

\textbf{Implementation Details.} SmolVLA-MoE is built on top of the SmolVLA architecture. The VLM backbone (SmolVLM2-500M) is frozen during training. The dense FFN layers of the action expert are replaced with MoE-FFN layers through Sparse Upcycling initialization. With 4 routed experts plus 1 shared expert, the total stored parameters of SmolVLA-MoE are $\sim$800M. During inference, only 1 expert FFN is activated per token, resulting in $\sim$100M activated parameters in the action expert module and $\sim$500M total activated parameters including the frozen VLM backbone---identical to the dense baseline SmolVLA. Key hyperparameters include: learning rate 1e-4 with AdamW optimizer, batch size 48, 120,000 training steps with 4,000 warmup steps, load balancing weight $\lambda=0.01$, expert initialization noise $\sigma=0.02$, and 10 Flow Matching denoising steps.

All experiments are conducted on a single NVIDIA RTX 4090 GPU (24GB VRAM). Each configuration is trained for 120,000 steps with the same random seed to ensure fair comparison.

\subsection{Main Results}
\label{sec:main_results}

Table~\ref{tab:main} presents the main comparison results of SmolVLA-MoE against multiple representative baseline methods across all four LIBERO task suites. We report the best configuration of SmolVLA-MoE (determined by the parameter study in Sect.~\ref{sec:ablation}).

\begin{table}[t]
  \caption{Main Results on the LIBERO Benchmark (Success Rate \%)}
  \label{tab:main}
  \centering
  \begin{tabular}{llccccc}
    \toprule
    Method & Params & Spatial & Object & Goal & LIBERO-10 & Avg. \\
    \midrule
    Diffusion Policy~\cite{chi2023diffusion_policy} & 321M & 78.3 & 92.5 & 68.3 & 50.5 & 72.4 \\
    Octo~\cite{octo2024} & 93M & 78.9 & 85.7 & 84.6 & 51.1 & 75.1 \\
    OpenVLA~\cite{kim2024openvla} & 7B & 84.7 & 88.4 & 79.2 & 53.7 & 76.5 \\
    $\pi_0$ (Paligemma)~\cite{black2024pi0} & 3B & 87 & 63 & 89 & 48 & 71.8 \\
    $\pi_0$~\cite{black2024pi0} & 3.3B & 90 & 86 & 95 & 73 & 86.0 \\
    SmolVLA~\cite{smolvla2025} & 500M & 87 & 93 & 88 & 63 & 82.75 \\
    \textbf{SmolVLA-MoE (Ours)} & \textbf{500M$^\dagger$} & \textbf{87} & \textbf{93} & \textbf{93} & \textbf{65} & \textbf{84.25} \\
    \bottomrule
  \end{tabular}
  \vspace{2pt}
\end{table}

As shown in Table~\ref{tab:main}, SmolVLA-MoE achieves an average success rate of 84.25\% with only $\sim$500M total activated parameters ($\sim$100M in the action expert module), significantly outperforming the much larger OpenVLA (7B, 76.5\%) and Octo (0.09B, 75.1\%), as well as Diffusion Policy (72.4\%) and $\pi_0$ (Paligemma-3B) (71.8\%). Compared to the direct baseline SmolVLA, SmolVLA-MoE improves the average success rate from 82.75\% to 84.25\% under the same activated parameter budget ($\sim$500M), a gain of 1.5 percentage points, indicating that the performance improvement comes from the expanded model capacity through sparse MoE rather than additional computational overhead.

Among all methods, only $\pi_0$ (3.3B) surpasses SmolVLA-MoE with an average success rate of 86.0\%, but its activated parameter count is approximately 3.3B---6.6$\times$ that of SmolVLA-MoE ($\sim$500M). This demonstrates the significant parameter efficiency advantage of SmolVLA-MoE, achieving performance close to large models at a fraction of the computational cost.

\textbf{Per-Suite Analysis.} SmolVLA-MoE matches SmolVLA on Spatial (87\%) and Object (93\%), while achieving a 5 percentage point improvement on Goal (88\% $\to$ 93\%) and a 2 percentage point gain on LIBERO-10 (63\% $\to$ 65\%). On Spatial and Object, the dense baseline already performs well---these tasks primarily require spatial understanding and object recognition, where a single FFN suffices. The most significant gains appear on Goal and LIBERO-10, which involve complex goal-state reasoning and multi-step planning; here, the MoE architecture enables specialized experts to handle distinct action generation strategies, providing a clear advantage. Notably, SmolVLA-MoE achieves the highest Object score (93\%) among all methods, surpassing even $\pi_0$ (3.3B, 86\%) by 7 points, while on LIBERO-10 the gap to $\pi_0$ (3.3B, 73\%) remains the largest, suggesting that compositional multi-step tasks may benefit from further scaling.

\textbf{Efficiency Analysis.} With $\sim$500M activated parameters, SmolVLA-MoE outperforms OpenVLA (7B) by 7.75 points and Diffusion Policy (321M) by 11.85 points, while trailing $\pi_0$ (3.3B) by only 1.75 points at 6.6$\times$ fewer activated parameters. The additional expert parameters ($\sim$300M) are only stored but not all activated during inference, providing ``free'' capacity expansion.

\begin{figure}[t]
  \centering
  \includegraphics[width=0.95\textwidth]{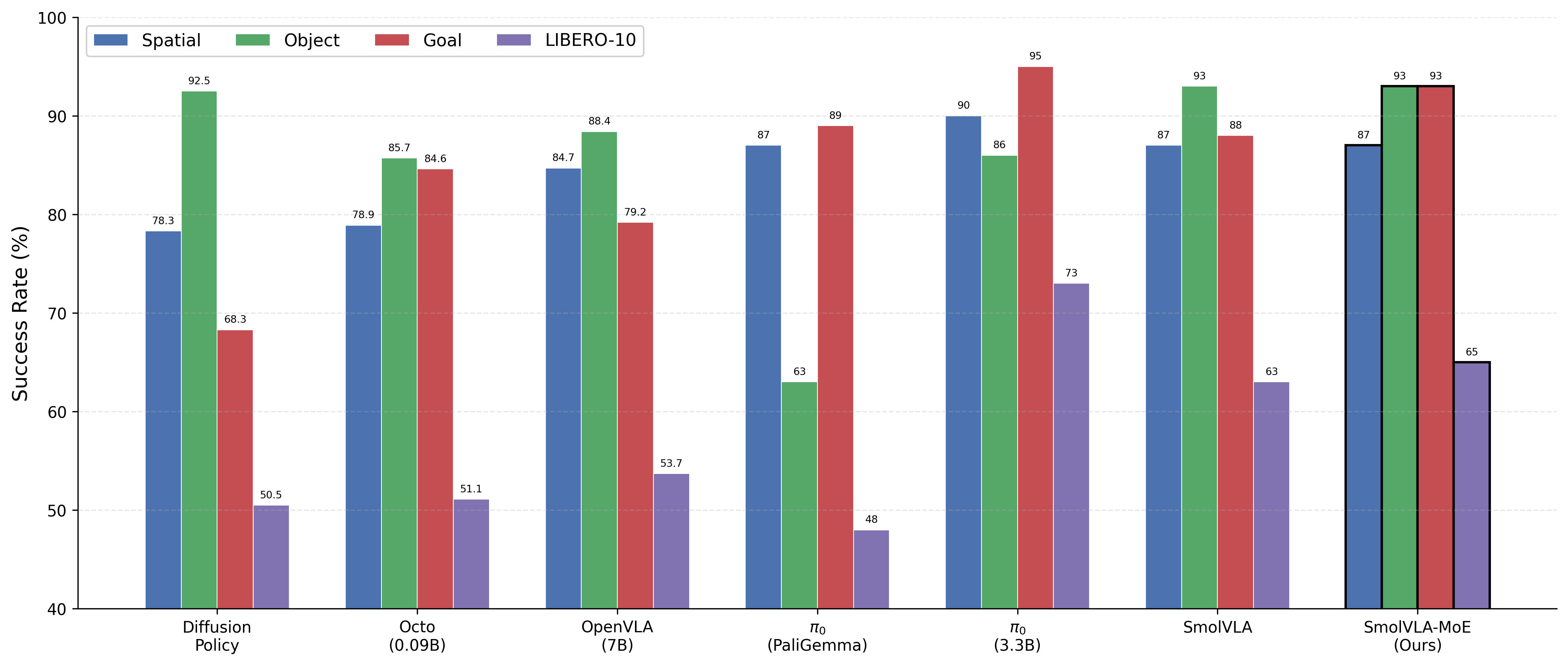}
  \caption{Per-suite success rate comparison across all methods on the LIBERO benchmark. SmolVLA-MoE (highlighted with black borders) achieves competitive or superior performance across all four suites with $\sim$500M total activated parameters ($\sim$100M in the action expert).}
  \label{fig:main_results}
\end{figure}

\subsection{Ablation Studies}
\label{sec:ablation}

We conduct systematic ablation studies to analyze the impact of two key design parameters: the number of activated experts per token (top-$K$) and the shared expert weight ($\alpha$). All ablation experiments use 4 experts with other hyperparameters fixed.

\subsubsection{Impact of Top-K and Shared Expert Weight.}

Table~\ref{tab:ablation_topk} presents the complete parameter study across all four LIBERO suites with different combinations of top-$K$ and shared expert weight $\alpha$.

\begin{table}[t]
  \caption{Ablation Study --- Top-K $\times$ Shared Expert Weight (Success Rate \%)}
  \label{tab:ablation_topk}
  \centering
  \begin{tabular}{cccccc}
    \toprule
    Top-K & $\alpha$ & Spatial & Object & Goal & LIBERO-10 \\
    \midrule
    1 & 0.2 & 86 & 91 & 89 & 62 \\
    1 & 0.3 & 87 & 93 & 93 & 65 \\
    2 & 0.2 & 77 & 89 & 89 & 64 \\
    2 & 0.3 & 78 & 95 & 90 & 60 \\
    \bottomrule
  \end{tabular}
\end{table}

\textbf{Top-K Analysis.} Top-1 routing outperforms top-2 on most suites, with the largest gap on Spatial (87\% vs. 78\% at $\alpha=0.3$, a 9-point difference). This suggests that with 4 experts, activating a second expert introduces redundant or conflicting signals rather than complementary information. The exception is LIBERO-10, where top-2 (64\% at $\alpha=0.2$) slightly outperforms top-1 (62\%), likely because multi-step compositional tasks benefit from combining multiple experts.

\textbf{Shared Expert Weight Analysis.} A higher $\alpha$ consistently improves performance. Under top-1 routing, increasing $\alpha$ from 0.2 to 0.3 yields gains across all suites, most notably on Goal (89\% $\to$ 93\%) and LIBERO-10 (62\% $\to$ 65\%). This indicates that preserving more pre-trained dense FFN knowledge through the shared expert provides a more stable foundation for routed expert specialization at the current training scale.

\begin{figure}[t]
  \centering
  \includegraphics[width=0.85\textwidth]{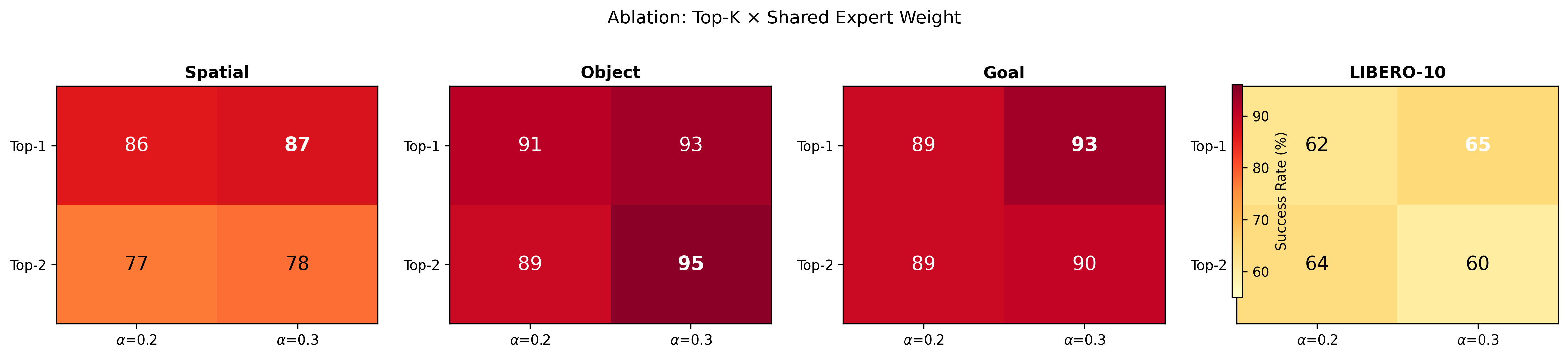}
  \caption{Heatmap visualization of the Top-K $\times$ shared expert weight ($\alpha$) ablation study across all four LIBERO suites. Best values in each suite are shown in bold. Top-1 routing with $\alpha=0.3$ achieves the best overall performance.}
  \label{fig:ablation_heatmap}
\end{figure}

\subsubsection{Impact of Shared Expert.}

To verify the necessity of the shared expert, we compare models with and without the shared expert component.

\begin{table}[t]
  \caption{Ablation Study --- Shared Expert (Success Rate \%)}
  \label{tab:ablation_shared}
  \centering
  \begin{tabular}{lc}
    \toprule
    Configuration & Goal \\
    \midrule
    SmolVLA-MoE (w/o shared expert, $\alpha = 0$) & 88 \\
    SmolVLA-MoE (w/ shared expert, $\alpha = 0.3$) & 93 \\
    \bottomrule
  \end{tabular}
\end{table}

Removing the shared expert ($\alpha=0$) causes the Goal success rate to drop from 93\% to 88\% ($-$5 points), validating its critical role. The shared expert preserves pre-trained dense FFN weights as a stable baseline, enabling routed experts to perform incremental task-specific specialization rather than learning all capabilities from scratch.

\subsubsection{Impact of Load Balancing Loss.}

We examine the impact of the load balancing loss by comparing training with and without it.

\begin{table}[t]
  \caption{Ablation Study --- Load Balancing Loss (Success Rate \%)}
  \label{tab:ablation_lb}
  \centering
  \begin{tabular}{lc}
    \toprule
    Configuration & Goal \\
    \midrule
    SmolVLA-MoE (w/o load balancing, $\lambda = 0$) & 90 \\
    SmolVLA-MoE (w/ load balancing, $\lambda = 0.01$) & 93 \\
    \bottomrule
  \end{tabular}
\end{table}

Removing the load balancing loss ($\lambda=0$) causes the Goal success rate to drop from 93\% to 90\% ($-$3 points). Without this regularization, the router converges to a locally optimal allocation where some experts are underutilized (expert collapse), preventing the model from fully leveraging the MoE capacity. A weight of $\lambda=0.01$ balances uniform utilization with routing flexibility.

\section{Conclusion}
\label{sec:conclusion}

In this paper, we propose SmolVLA-MoE, a framework that introduces FFN-level sparse Mixture-of-Experts into the action expert of Vision-Language-Action models for multi-task robotic manipulation. By replacing dense FFN layers with MoE-FFN layers containing multiple parallel expert FFNs and a learnable token-level router, SmolVLA-MoE enables each token to adaptively select specialized experts for action generation, scaling model capacity without increasing inference cost.

Experiments on the LIBERO benchmark demonstrate that SmolVLA-MoE achieves superior performance over the SmolVLA baseline on all four task suites (Spatial, Object, Goal, and LIBERO-10), while maintaining the same total activated parameter count of $\sim$500M as the dense baseline (with only $\sim$100M activated in the action expert module). Systematic ablation studies reveal that top-1 routing outperforms top-2 routing on tasks requiring precise action generation, indicating that activating fewer but more specialized experts is preferable when expert differentiation is limited. The shared expert weight provides an effective control knob for balancing stability and specialization across different task types.

\textbf{Limitations and Future Work.} Our evaluation is limited to the LIBERO benchmark in simulation. Future work includes extending to real-world tasks and other benchmarks (e.g., MetaWorld~\cite{yu2020metaworld}, RLBench~\cite{james2020rlbench}), exploring layer-wise adaptive expert configurations, and combining MoE with complementary efficiency techniques such as quantization and knowledge distillation.

\begin{credits}
\subsubsection{\ackname}
This work was supported by the Fundamental Research Funds for the Central Universities (Grant No. 2662025ZKPY009).

\subsubsection{Declaration on Generative AI}
We used Claude as a writing assistant to improve the language quality and readability of this paper. The AI tool was employed solely for refining sentence structure, enhancing clarity, and polishing academic writing style. All technical content, experimental results, and scientific contributions are entirely original work by the authors.
\end{credits}

\bibliographystyle{splncs04}
\bibliography{bibliography}

\end{document}